%% file: main.tex
\documentclass{article}
\PassOptionsToPackage{square,sort,numbers, comma}{natbib}

\usepackage[preprint]{neurips_2024}

\usepackage{hyperref}
\hypersetup{colorlinks,allcolors=black}

\usepackage[utf8]{inputenc} 
\usepackage[T1]{fontenc}    
\usepackage{hyperref}       
\usepackage{url}            
\usepackage{booktabs}       
\usepackage{amsfonts}       
\usepackage{nicefrac}       
\usepackage{microtype}      
\usepackage[table]{xcolor}  

\usepackage{alltt}
\usepackage{amsmath}
\usepackage{amssymb}
\usepackage{array}
\usepackage{arydshln}
\usepackage{authblk}
\usepackage{bm}
\usepackage{booktabs}
\usepackage{colortbl}
\usepackage{enumitem} %
\usepackage{graphicx}
\usepackage{epstopdf}
\usepackage[mathscr]{euscript}
\usepackage{float}
\usepackage{geometry}

\usepackage{lineno}         
\usepackage{lipsum}
\usepackage{marvosym}
\usepackage{mathtools} 
\usepackage{multirow}
\usepackage{makecell}
\usepackage{pifont} 
\newcommand{\xmark}{\ding{55}} 
\usepackage{subcaption}
\usepackage{tabularray}
\usepackage{tabularx}

\title{TAMBRIDGE: Bridging Frame-Centered Tracking and 3D Gaussian Splatting for Enhanced SLAM}

\author{
    Peifeng Jiang\textsuperscript{1}, Hong Liu\textsuperscript{1,\Letter},
    Xia Li\textsuperscript{2,\Letter}, Ti Wang\textsuperscript{1}, Fabian Zhang\textsuperscript{2}, Joachim Buhmann\textsuperscript{2} \\
  \textsuperscript{1}National Key Laboratory of General Artificial Intelligence, \\ Peking University, Shenzhen Graduate School \\
  \textsuperscript{2}Institute for Machine Learning, Department of Computer Science, ETH Zurich \\
  \texttt{jpf@stu.pku.edu.cn, hongliu@pku.edu.cn, xia.li@inf.ethz.ch} \\ \texttt{tiwang@stu.pku.edu.cn, fabzhang@ethz.ch, jbuhmann@inf.ethz.ch} \\
}

\begin{document}

\maketitle
\input{00_abstract}
\input{01_introduction}
\input{02_related_work_pre}
\input{03_method}
\input{04_experiment}
\input{05_conclusion}
\begin{ack}

This project is supported by the National Natural Science Foundation of China (No.62373009) and the interdisciplinary doctoral grant (iDoc 2021-360) from the Personalized Health and Related Technologies (PHRT) of the ETH domain, Switzerland.
\end{ack}

\bibliographystyle{plain}
\bibliography{IEEE}

\clearpage
\input{appendix}

\end{document}

%% file: 00_abstract.tex

\vspace{-4mm}
\begin{figure}[htbp]
    \centering
    \includegraphics[width=0.7\linewidth]{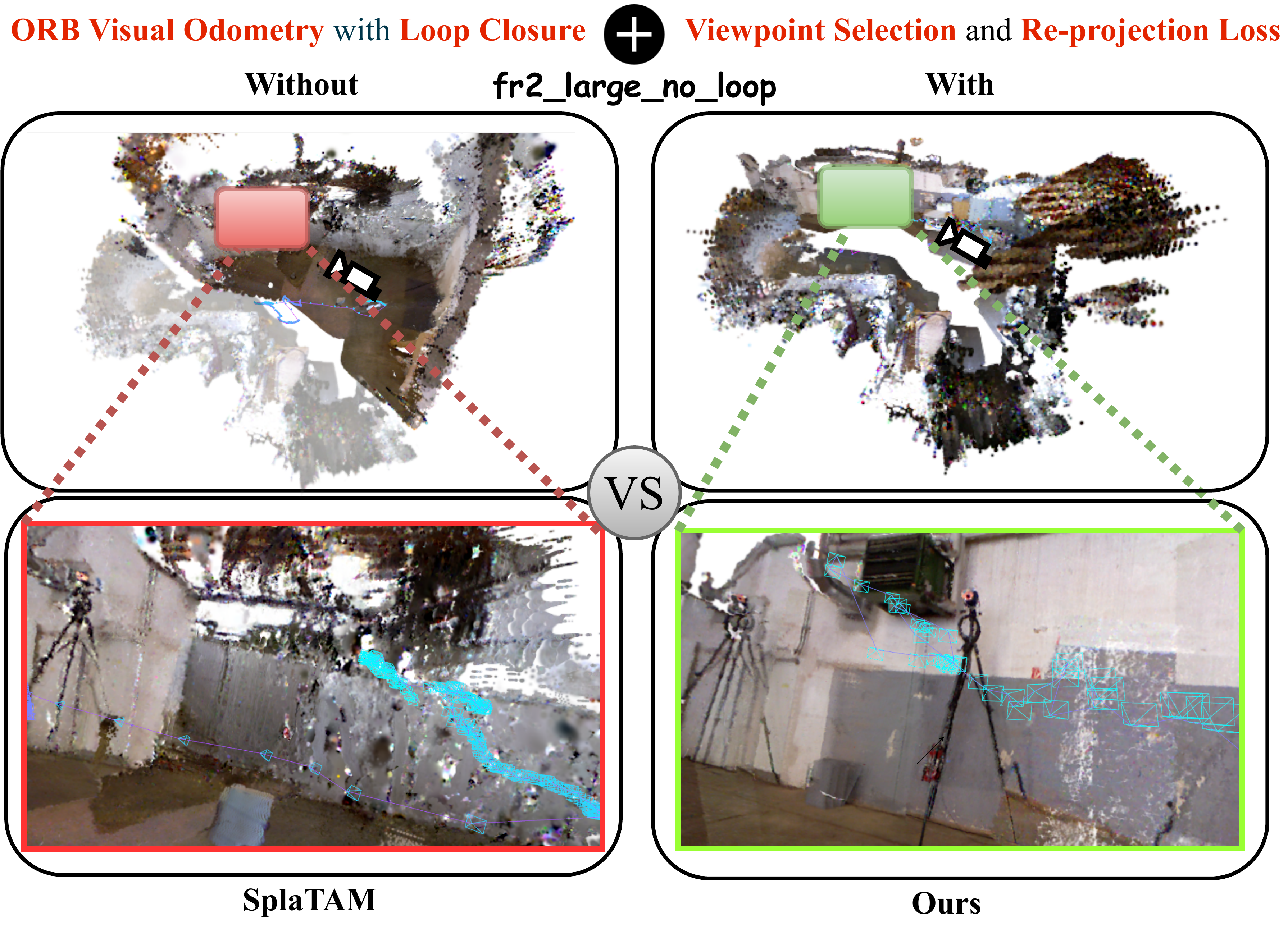}
    \caption{By seamlessly integrating ORB Visual Odometry with viewpoint selection and re-projection loss, our method significantly improves the robustness towards sensor noise and motion blur especially in long-session robotic tasks.}
    \label{fig:projection}
    \vspace{-2mm}
\end{figure}

\begin{abstract}
The limited robustness of 3D Gaussian Splatting (3DGS) to motion blur and camera noise, along with its poor real-time performance, restricts its application in robotic SLAM tasks. Upon analysis, the primary causes of these issues are the density of views with motion blur and the cumulative errors in dense pose estimation from calculating losses based on noisy original images and rendering results, which increase the difficulty of 3DGS rendering convergence. Thus, a cutting-edge 3DGS-based SLAM system is introduced, leveraging the efficiency and flexibility of 3DGS to achieve real-time performance while remaining robust against sensor noise, motion blur, and the challenges posed by long-session SLAM. Central to this approach is the Fusion Bridge module, which seamlessly integrates tracking-centered ORB Visual Odometry with mapping-centered online 3DGS. Precise pose initialization is enabled by this module through joint optimization of re-projection and rendering loss, as well as strategic view selection, enhancing rendering convergence in large-scale scenes. Extensive experiments demonstrate state-of-the-art rendering quality and localization accuracy, positioning this system as a promising solution for real-world robotics applications that require stable, near-real-time performance. Our project is available at \url{https://ZeldaFromHeaven.github.io/TAMBRIDGE/}
\end{abstract}

%% file: 01_introduction.tex
\section{Introduction}
\label{intro}

Without prior knowledge, Real-time Simultaneous Localization and Mapping (SLAM) is essential for intelligent robots to form a comprehensive environmental understanding of the environment. Over the last two decades, the primary challenge in SLAM has been to balance computational demands between tracking accuracy and mapping fidelity. Frame-centered, sparse, and effective information can improve localization accuracy but lacks a high-level understanding of the entire scene. Conversely, mapping-centered, dense, and global information contributes to building a globally consistent map. Thus, the essence of the challenge lies in the contradiction between these two aspects: efficiency but sparse, and dense but noisy.


Methods prioritizing localization accuracy are known as "tracking-centered" methods, while those focusing on high-fidelity mapping are called "mapping-centered" methods. Initially, due to computational limitations, tracking-centered methods became a focal area of research since their high real-time performance, accuracy, and robustness in localization, including the use of feature point clouds~\cite{davison2007monoslam, orb-slam,orb-slam2}, surfels \cite{whelan2015elasticfusion,schops2019bad}, depth maps \cite{stuhmer2010real,newcombe2011dtam} and implicit representations \cite{newcombe2011kinectfusion,niessner2013real,dai2017bundlefusion}.
However, these methods often lack the capacity for consistent scene understanding.


As computing capabilities evolve, the convergence issues in mapping-centered methods are gradually addressed through judicious selection and processing of scene representations. Concurrently, the operational efficiency of these paradigms is also steadily improving. A significant advancement is adopting neural radiance fields (NeRF)~\cite{mildenhall2021nerf}, which utilizes implicit neural networks for globally consistent representations and incorporates ray-cast triangulation for high-fidelity rendering and novel view synthesis. Nevertheless, the extensive computational demands of neural network inference and dense, per-pixel optimization calculations challenge the robustness of NeRF-based SLAM methods against sensor noise and motion blur in real-world robotic scenarios.

Recently, the 3D Gaussian Splatting (3DGS)~\cite{kerbl20233dgaussiansplatting} utilizes Gaussian primitives that are both interpretable and editable, serving as a globally consistent representation of the scene. This method facilitates even more rapid, differentiable rasterization rendering and generates higher-fidelity novel views through Gaussian splatting. SLAM systems incorporating 3DGS now achieve localization accuracy comparable to NeRF-based SLAM while significantly outperforming in terms of rendering quality and speed. However, these systems still face persistent challenges typical of mapping-centered SLAM, including non-real-time convergence rates and high sensitivity to random noise, which have difficulties in handling long-term, high-noise robotic tasks. 

To solve this problem, our focus is to enhance the convergence capabilities of 3DGS towards online rendering. Random noise of real sensors and increasing perspective overlap are two main barriers to the rendering convergence speed, especially in long-session robotic tasks. Therefore, locally optimal poses from tracking-centered SLAM are leveraged to provide a robust initial estimate. By jointly optimizing sparse re-projection and dense rendering errors, the impact of image noise is effectively reduced. Additionally, a viewpoint selection strategy is adapted to effectively pick viewpoints suitable for online reconstruction, ensuring viewpoint sparsity and reducing overlap. The above two processes are integrated into a plug-and-play Fusion Bridge module, which acts as an intermediary between the tracking-centered frontend (ORB~\cite{rublee2011orb} Visual Odometry) and the mapping-centered backend (Online 3DGS).
As shown in Figure \ref{fig:projection}, our method obtains better tracking accuracy for long-session tasks by using the ORB Visual Odometry's loop closure to reduce the accumulated trajectory error. The introduction of viewpoint selection and re-projection loss in joint optimization helps us to eliminate the view overlapping problem and enhance the robustness towards sensor noise, resulting in better rendering performance.
Extensive experiments on real-world datasets show that our method achieves state-of-the-art (SOTA) rendering quality and localization accuracy in long-session robotic tasks. It is also the first 3DGS-based SLAM system capable of consistently achieving (5+ FPS) real-time performance. Our contributions can be summarized as follows:
\begin{itemize}
    \item The significance of precise pose initialization and strategic viewpoint selection in facilitating the convergence of online 3DGS, especially in large-session tasks, is emphasized.
    \item A plug-and-play Fusion Bridge module is introduced, seamlessly integrating tracking-centered SLAM frontend with online 3DGS-centered backend.
    \item A SLAM system is developed that consistently delivers stable, near real-time performance (>5 FPS) for long-session robotic tasks.
\end{itemize}

%% file: 02_related_work_pre.tex
\section{Related Works}

\subsection{Tracking-centered and Mapping-centered SLAM}
Tracking-centered SLAM has evolved into three primary approaches: feature-based, deep learning-based, and hybrid. Feature-based approaches like MonoSLAM \cite{davison2007monoslam}, PTAM \cite{klein2007parallel}, and ORB-SLAM \cite{orb-slam, orb-slam2} utilize traditional strategies such as extended Kalman Filters~\cite{hoshiya1984structural} and place recognition to estimate camera motion real-time, with ORB-SLAM standing out among feature-based methods for its effective integration of tracking, local mapping, and loop closing, ensuring real-time performance. Deep learning-based methods, such as DeepVO \cite{deep-vo, dosovitskiy2015flownet} and UnDeepVO \cite{undeep-vo}, employ neural networks to directly estimate camera poses and environmental depth, offering innovative solutions but facing accuracy challenges. Hybrid approaches like DF-SLAM \cite{kang2019df, orb-slam2} and SuperPointVO \cite{detone2018self, detone2018superpoint} merge deep learning for feature extraction with traditional frameworks, aiming to bridge the gap between conventional methods and modern deep learning innovations. These hybrid systems show promise in improving the accuracy and robustness of SLAM, making them a significant development in the field.

The mapping-centered SLAM paradigm emphasizes the util of 3D representation within the framework and leverages global information through consistently reconstructed 3D maps from tracking data. A prominent technology in this domain is the NeRF, which allows for high-fidelity view-synthesis through the implicit representation based on neural networks. A growing body of dense neural SLAM methods \cite{Sucar:etal:ICCV2021,Zhu2022CVPR,johari2023eslam,wang2023coslam,xu2022point,go-slam} has been developed, which build a high-fidelity map and achieve the same level of tracking accuracy compared with tracking-centered works.

\subsection{NeRF based SLAM}
The pursuit of photorealistic scene capture has popularized differentiable volumetric rendering, notably through techniques like NeRF. These methods use a single multilayer perceptron (MLP) to derive opacity and color along pixel rays, optimized via multiview information. Due to the high computational costs of training these models, alternatives like multi-resolution voxel grids \cite{fridovich2022plenoxels,Zhu2022CVPR,liu2020neural} and hashing functions \cite{mueller2022instant} have been developed. Techniques such as Point-NeRF \cite{xu2022point} expedite 3D reconstruction using neural point clouds and feature interpolation, maintaining high rendering quality without relying on MLPs. The evolution of differentiable volumetric rendering highlights the significance of developing efficient rendering techniques to facilitate high-fidelity 3D reconstructions in fields like robotics and computer vision, by moving beyond traditional MLP-based methods.
\subsection{3DGS based SLAM}
Utilizing 3D Gaussian Splatting (3DGS) for scene representation, initiatives like SplaTAM \cite{keetha2023splatam}, MonoGS \cite{matsuki2023gaussian}, GS-SLAM \cite{Yan2023GSSLAMDV}, and Photo-SLAM \cite{huang2023photo} achieve efficient, deformable, and modifiable geometry. These methods opt for Gaussian splatting over ray marching as in NeRF, resulting in faster, high-fidelity rendering. SplaTAM employs isotropic 3D Gaussians for scene reconstruction, facilitating rapid differentiable rasterization. It segments the SLAM process into localization, densification, and mapping, achieving high localization accuracy and superior reconstruction capabilities. However, its localization phase is notably sensitive to motion blur, depth noise, and aggressive rotations, which along with significant time overheads, limits its application in robotics. Similarly, MonoGS uses a comparable framework with enhanced features like analytic Jacobian determinants and Gaussian shape regularization for camera pose estimation, extending input capabilities from RGBD to monocular systems. Although it surpasses NeRF-based SLAM in reconstruction performance, it shares similar sensitivities to motion conditions and exhibits increased time overhead due to per-pixel photometric and geometric residuals, impacting robustness to noise.

%% file: 03_method.tex
\section{Method}
\label{method}
\begin{figure}[htbp]
        \centering        \includegraphics[width=1.0\linewidth]{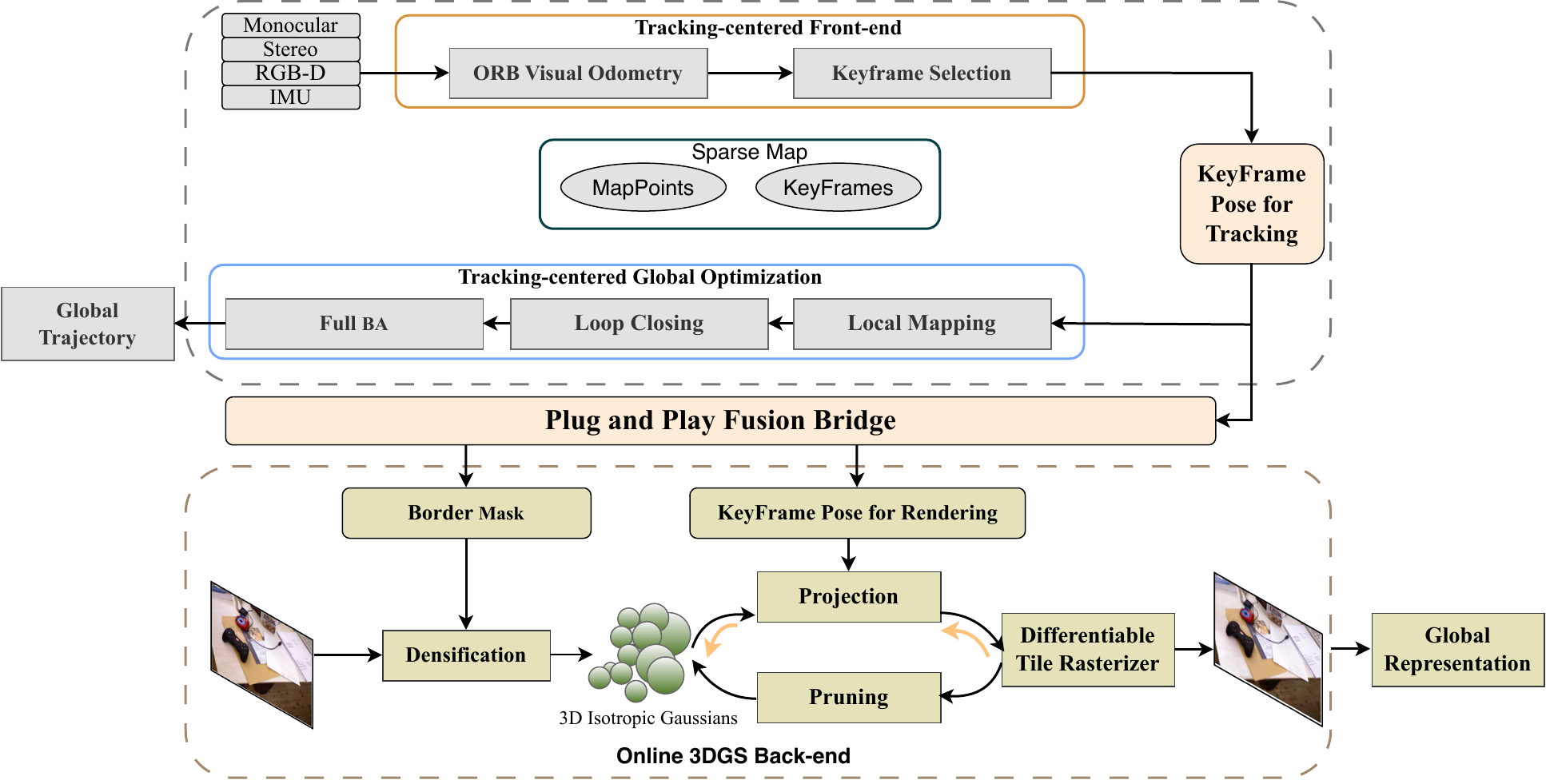}
\caption{Overview. ORB-based visual odometry analyzes RGB-D data to estimate poses and select keyframes for local mapping. These keyframes feed into the Global Optimization featuring Bundle Adjustment (BA) to refine the path. Simultaneously, the Fusion Bridge module selects reconstruction frames and calculates rendering poses and Border Masks. An online 3DGS backend processes the selected frames to create a globally consistent, high-fidelity scene representation.}
\label{fig:overview}  
\end{figure}
\vspace{-4mm}

To address convergence issues in 3DGS-based SLAM systems caused by sensor noise, motion blur, and overlapping views as shown in Figure \ref{fig:overview}, we propose the following modifications: 
\begin{itemize}
    \item \textbf{The Tracking-centered Frontend Module} obtains initial pose estimates and calculates keyframe sequences using an ORB visual odometry module, similar to ORB-SLAM3.
    
    \item \textbf{The Tracking-centered Global Optimization Module} optimizes the final trajectory through global Bundle Adjustment (BA) after local mapping and loop closing.
    
    \item \textbf{The Plug and Play Fusion Bridge Module} utilizes covisibility to select reconstruction frames from the keyframe sequence further and jointly optimizes rendering poses and border masks by minimizing re-projection errors and color-depth rendering errors.
    
    \item \textbf{The Online 3DGS Backend Module} uses the poses of reconstruction frames and border masks to construct a high-fidelity, globally consistent Gaussian scene representation nearly in real-time.
\end{itemize}

The coarse-to-fine pose estimation during the Tracking-centered frontend and Global Optimization is described in Section \ref{sec:3.2}. Section \ref{sec:3.3} will elaborate on the Plug and Play Fusion Bridge module, and Section \ref{sec:3.4} will discuss the online 3DGS reconstruction.
\subsection{Pose Estimation}
\label{sec:3.2}


In the Tracking-centered Frontend and Global Optimization process, pose is refined from coarse to fine based on a sparse feature point cloud based on Bundle Adjustment (BA) \cite{triggs2000bundle}.

In the Tracking-centered Frontend, re-projection error, used for short-term data associations, is calculated from the matched ORB feature points extracted from consecutive frames as follows:
\begin{equation}
\label{lba}
    L_{r p j}=\min_{\{[R|t]_j\}} \sum_{i j}\left\|u_{i j}-\pi\left(\mathscr{C}_{j}, P_{i}\right)\right\|_{2}.\
\end{equation}
The projection \( \pi(\mathscr{C}_{j}, P_i) \) converts a 3D point $ P_i = [X, Y, Z]^T $ to pixel position \( u_{ij} \)  using intrinsic parameters \( K_j \) and extrinsic parameters \( [R|t]_j \), which include rotation \( R_j \) and translation \( t_j \) of camera \( \mathscr{C}_{j} \). The process optimizes camera poses \( \{[R|t]_j\} \) by minimizing the reprojection error \( L_{rpj} \).

During global optimization, new map points are added to the local map by triangulation. The reprojection error for both camera poses of keyframes (described by \(K\) and \([R|t]\)) and the observed map points is minimized through full BA, detailed in equation~\ref{for:ba}. This process constitutes a nonlinear least squares problem, tackled using the Levenberg-Marquardt method according to~\cite{more2006levenberg}.
\begin{equation}
\label{for:ba}
    L_{r p j}=\min_{\{[R|t]_j\},{P_i}} \sum_{i j}\left\|u_{i j}-\pi\left(\mathscr{C}_{j}, P_{i}\right)\right\|_{2}.\
\end{equation}
\subsection{Plug and Play Fusion Bridge}
\label{sec:3.3}
\begin{figure}
    \centering
    \includegraphics[width=1.0\linewidth]{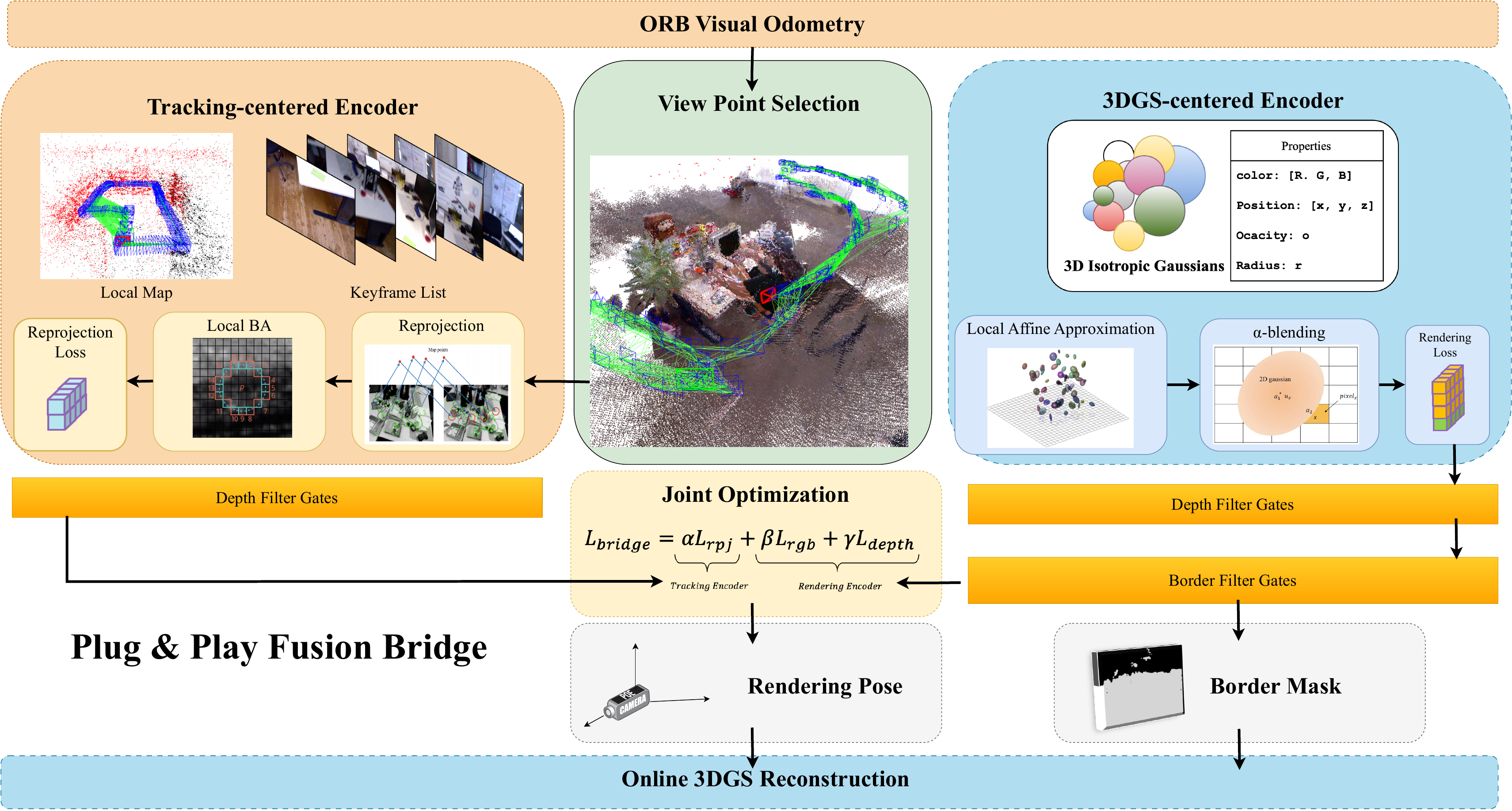}
    \caption{
The Fusion Bridge module selects reconstruction keyframes from a local map based on viewpoint covisibility. It projects the 3DGS and local map point cloud onto the reconstruction frame, filtering projections through pixel gates. The module then optimizes the pose by jointly minimizing rendering and point cloud reprojection losses, setting the initial rendering pose for the Online 3DGS.}
    \label{fig:fusion_process}
\end{figure}
\vspace{-2mm}
The fusion bridge process shown in Figure \ref{fig:fusion_process} consists of five components: \textbf{viewpoint selection}, a\textbf{ Tracking-centered encoder}, a \textbf{3DGS-centered encoder}, \textbf{ filter gates}, and a \textbf{joint optimization module}. This process aims to minimize the gap between the sparse point cloud generated by visual odometry and the backend used for Gaussian Splatting, specifically 3DGS with selected viewpoints to get proper and sparse poses for reconstruction.

\textbf{The View Point Selection} involves selecting advantageous viewpoints for rendering by filtering through the covisibility relationships among keyframes. The covisibility between two frames should fall within a closed interval, balancing sufficient matching points for calculating reprojection errors and maintaining adequate diversity between viewpoints: 
\begin{equation}
R_i = \begin{cases} 
1 & \text{if } \beta \leq M_{i-1, i} < \alpha \cdot T_{i-1, i} \\
0 & \text{otherwise}
\end{cases},
\end{equation}
where \( R_i \) indicates if the current keyframe \( i \) is selected for reconstruction (1 for yes, 0 for no). \( M_{i-1, i} \) counts the matched feature points. \( T_{i-1, i} \) is the total number of matches in the previous keyframe, which, alongside \( \alpha \), sets the percentage threshold for maximum matches, while \( \beta \) specifies the minimum required matches for selection.

\textbf{Tracking-centered Encoder} encodes the sparse point cloud generated by visual odometry, enhancing the anchoring effect of robust feature-based tracking poses. By re-projecting the visible map points from the local map of the rendering frame, the reprojection loss can be optimized as Equation \ref{lba}.

\textbf{3DGS-centered Encoder} encodes the 3DGS used for Gaussian Splatting, where local affine transformations project these isotropic 3D Gaussians onto the camera plane, and the alpha-blending algorithm is then used to render the color and depth image, obtaining the rendering loss \(L_{rgb}\) and \(L_{depth}\). Additionally, the border masks \(M_b(\mathbf{p})\) that reflect the uncertainty of rendered pixels \(\mathbf{p}\) are computed based on the opacity information of the Gaussians as follows:
\begin{equation}
 M_b(\mathbf{p}) = \sum_{i=1}^{n} f_i(\mathbf{p}) \prod_{j=1}^{i-1} (1 - f_j(\mathbf{p})).
\end{equation}
\textbf{Filter Gates} eliminate noise during the optimization. The depth gates filter out depth anomalies from point clouds and Gaussians. The border gate filters out rendered pixels with high uncertainty according to the border mask:
\begin{equation}
G(\mathbf{p}) = \begin{cases} 
1 & \text{if } D_{c}(\mathbf{p})>0, D_{g}(\mathbf{p})>0\text{ and } M_b(\mathbf{p}) > \gamma \\
0 & \text{otherwise}
\end{cases},
\end{equation}
where \( G(\mathbf{p}) \) indicates if the current pixel \( \mathbf{p} \) is selected for optimization (1 for yes, 0 for no).  \(D_{c}(\mathbf{p})\) and \(D_{g}(\mathbf{p})\) is the depth of cloud point and Gaussian at pixel \( \mathbf{p} \). $\gamma$ is the threshold of the minimize border mask value of \( \mathbf{p} \). 

\textbf{Joint Optimization}
employs alternating optimization based on the rendering error of depth and color, as well as the reprojection error based on feature coordinates:
\begin{equation}
\mathbf{L}=w_1 \cdot \mathbf{L}_{rpj}+ w_2 \cdot \mathbf{L}_{c} + w_3 \cdot \mathbf{L}_{d}.
\end{equation}
Here, $\mathbf{L}$ denotes the aggregate loss, incorporating the reprojection loss of the point cloud features, $ \mathbf{L}_{rpj}$, the rendering loss of the Gaussian color $ \mathbf{L}_{c}$ and depth $ \mathbf{L}_{d}$. The coefficients $w_1$, $w_2$ and $ w_3$ respectively represent the weight hyper-parameters for these three types of loss. 
\subsection{Online 3DGS Reconstruction}
\label{sec:3.4}
The online 3DGS reconstruction method enables near real-time reconstruction using sparse frames and border masks filtered through the Fusion Bridge module. Differing from primary 3DGS works, SLAM operates without global pose and point cloud priors,  requiring incremental reconstruction based on the current and preceding frames. Presently, 3DGS-based SLAM typically utilizes incremental densification, initializing under-reconstructed pixels in the current frame as new 3DGS, marking the key distinction between online and offline 3DGS.

\textbf{Gaussian Representation.} Similar to SplaTAM, we simplified the model from~\cite{kerbl20233dgaussiansplatting} by using view-independent colors and isotropic Gaussians. Each Gaussian is defined by eight parameters: three for RGB color $\bf{c}$, three for center position $\bm{\mu} \in \mathbb{R}^3$, one for radius $r$, and one for opacity $o \in [0,1]$, as illustrated in Figure~\ref{fig:fusion_process}. Each Gaussian affects a point \( \mathbf{x} \) in 3D space, belonging to \( \mathbb{R}^3 \), based on the standard (unnormalized) Gaussian equation, with the influence modulated by its opacity.
\begin{align}
    f(\mathbf{x}) = o \exp\left(-\frac{\|\mathbf{x} - \boldsymbol{\mu}\|^2}{2r^2}\right).\label{eq:gauss}
\end{align}
\textbf{Densification.} Pixels outside the Border Mask are under-reconstructed, lacking a robust Gaussian representation, necessitating densification. During the first frame, all pixels from that frame are selected for reconstruction. During the reconstruction, the new Gaussians are added with a radius of one pixel length on the pixel plane as shown in equation~\ref{d}, where \(f\) represents the focal length.
\begin{equation}
    \label{d} d = rf.
\end{equation}
\textbf{Projection.} Global Gaussians are projected onto the rendering pose plane via local affine transformations, resulting in a splatted 2D projection of each Gaussian.

\textbf{Differentiable Rasterization.} Gaussian Splatting involves rendering an RGB image by first arranging a set of 3D Gaussians in a front-to-back order relative to the camera pose. The RGB image is then efficiently generated by alpha-blending the 2D projections of these splatted Gaussians in sequential order within the pixel space. The color of a rendered pixel ${\bf p} = (u,v)$ is defined by the equation:
\begin{equation}
    C(\mathbf{p}) = \sum_{i=1}^{n} \mathbf{c}_i f_i(\mathbf{p}) \prod_{j=1}^{i-1} (1 - f_j(\mathbf{p})),
\end{equation}
where each $f_i(\bf{p})$ is calculated using Equation~\ref{eq:gauss}. The specific equations for transforming the Gaussians to 2D pixel-space are:
\begin{equation}
        \boldsymbol{\mu}^{\textrm{2D}} = K \frac{E_t \boldsymbol{\mu}}{d}, \quad r^{\textrm{2D}} = \frac{f r}{d}, 
\end{equation}
Here, $K$ represents the camera's intrinsic matrix, $E_t$ is the extrinsic matrix indicating the camera's rotation and translation at frame $t$, $f$ is the known focal length, and $d$ is the depth of each Gaussian in camera coordinates which is defined as follows:
\begin{equation}
    D(\mathbf{p}) = \sum_{i=1}^{n} d_i f_i(\mathbf{p}) \prod_{j=1}^{i-1} (1 - f_j(\mathbf{p})),
\end{equation}
The Gaussian parameters are iteratively refined using gradient-based optimization by minimizing the specified loss, while the pose of the camera is kept constant:
\begin{equation}
\mathbf{L}_m=(1-\zeta)(w_4\mathbf{L}_1(D(\mathbf{p})+w_5\mathbf{L}_1(C(\mathbf{p}))+\zeta\mathbf{L}_{SSIM}.
\end{equation}
This is an L1 loss on both the depth and color renders, with different weights \(w_4\) and \(w_5\). Furthermore, we add an SSIM loss to RGB rendering.

\textbf{Gaussian Pruning.} After updating the Gaussian parameters, Gaussians with opacity below zero or excessively high are removed during rendering, as they are deemed redundant or non-contributory to the reconstruction:
\begin{equation}
P(\mathbf{p}) = \begin{cases} 
0 & \text{if } 0 \leq o < \tau \\
1 & \text{otherwise}
\end{cases},
\end{equation}
where \(P(\mathbf{p})\) indicates if the pixel \(\mathbf{p}\) is pruned (1 for yes, 0 for no) and \(\tau\) is the threshold of opacity.

%% file: 04_experiment.tex
\section{Experiments}
\label{experiment}
\subsection{Experimental Setup}

\textbf{Implementation Details}. Experiments were performed with an Intel i7-9700 CPU and an NVIDIA RTX 3090 GPU. We developed the TAMBRIDGE code in Python, C++, and CUDA, computed reprojection errors using the G2O framework with the Levenberg-Marquardt algorithm and used the differentiable rasterization modules from the paper~\cite{kerbl20233dgaussiansplatting}. 

\textbf{Dataset and Baselines.} 
The TUM RGB-D dataset ~\cite{sturm2012benchmark}, collected by Pioneer robots and handheld devices, is characterized by sensor noise, motion blur, and aggressive rotations. These 18 sequences were categorized into four types: common, long-session mobile, handheld, and tiny-motion. For the common type including 3 sequences, we conducted comparisons with baseline methods of NeRF-based SLAM, SplaTAM, the SOTA of 3DGS-based SLAM, and ORB-SLAM3. Comparisons were also made with SplaTAM in the other three categories including 15 sequences,  since most NeRF-based methods didn't share settings on these sequences. 

\textbf{Evaluation Metrics.} For the precision of the localization, we calculated the Root Mean Square Error (RMSE) between the ground truth and the estimated trajectories. For online 3DGS reconstruction, we further adapt the peak signal-to-noise ratio (PSNR), SSIM, and LPIPS by following~\cite{keetha2023splatam} to evaluate the rendering performance of each frame. For real-time performance, we measure the frame rate (FPS) of the whole system.

\subsection{Evaluation}
\input{tum_rendering}
\textbf{Evaluation on Common Sequences}. On these three sequences commonly used in SLAM domain, we compare two versions of our method with different FPS, as shown in Table \ref{tab:tum_rendering}. Towards the version of steadily achieve \textbf{5 FPS}, our localization performance significantly surpasses, \textbf{at least} \textbf{3 times}, than other methods, matching the level of tracking-based SLAM like ORB-SLAM3. In terms of rendering quality, our method \textbf{greatly outperforms} other NeRF-based SLAM methods and is \textbf{on par }with SplaTAM. Regarding real-time performance, our method achieves a \textbf{near real-time} performance (5 FPS) for optimal localization and rendering that is more than \textbf{6 times} the average frame rate of SplaTAM and on average {\textbf{50 times}} over other NeRF-based methods. Towards the version with an average \textbf{10 FPS+} frame rate with the same tracking accuracy, the rendering quality of our method is still \textbf{on par with} the SOTA NeRF-based method with on average over \textbf{100 times} faster than these methods, achieving \textbf{real-time} performance.

\textbf{Evaluation on Long-Session SLAM Sequences}. Towards robotic long-session SLAM tasks that typically involve larger scenes and longer task durations, our method \textbf{consistently outperforms} SplaTAM in terms of localization accuracy and rendering quality. As shown in Table \ref{tab:tum_mobile}, SplaTAM exhibits significant localization errors and inferior reconstruction quality over long distances. Our methods, which steadily achieve \textbf{5 FPS} near real-time performance, by employing viewpoint filtering and robust initial estimates from joint optimization of reprojection errors, enables the algorithm to converge to a better rendering performance \textbf{20 times faster} than SplaTAM in long-duration and long-distance rendering tasks.

\input{tum_mobile}

\textbf{Evaluation on Handheld Sequences}.
Handheld sequences, taken by a handheld camera for extended periods, underscore the other methods vulnerability to motion blur. From Table~\ref{tab:tum_handheld}, SplaTAM fails at \textbf{fr2/hemi} and \textbf{fr2/kidn} because of its extended periods and great motion blur, while our method can achieve consistently accurate tracking and high-fidelity rendering with an average PSNR over than \textbf{20} in a near real-time performance (5 FPS).
\input{tum_handheld}
\vspace{-4mm}
%
%
\begin{table}[htbp]
    \centering
    \footnotesize
    \setlength{\tabcolsep}{1.8mm}
    \caption{\textbf{Experimental Results without Fusion Bridge.} }
    \begin{tabular}{cccccccccc}
    \toprule
\multicolumn{1}{c}{Common Sequences}&  \multicolumn{3}{c}{fr1-desk}& \multicolumn{3}{c}{fr2-xyz}& \multicolumn{3}{c}{fr3-office}\\\cmidrule(lr){1-1} \cmidrule(lr){2-4} \cmidrule(lr){5-7}\cmidrule(lr){8-10}
{\scriptsize Method} &{\scriptsize PSNR $\uparrow$} &{\scriptsize SSIM $\uparrow$} & {\scriptsize LPIPS $\downarrow$}& {\scriptsize PSNR $\uparrow$} &{\scriptsize SSIM $\uparrow$} & {\scriptsize LPIPS $\downarrow$}& {\scriptsize PSNR $\uparrow$} &{\scriptsize SSIM $\uparrow$ }& {\scriptsize LPIPS $\downarrow$} \\
    \midrule
\textbf{Ours (with bridge)}&\textbf{21.22}&\textbf{0.88}&\textbf{0.19}&\textbf{23.44}&\textbf{0.90}&\textbf{0.10}&\textbf{20.15}&\textbf{0.82}&\textbf{0.25}\\ 
    \textbf{Ours (no bridge)}&13.15&0.45&0.48&14.59&0.59&0.35&15.59&0.53&0.46\\ 
    \bottomrule
    \end{tabular}
    \label{tab:fusion}
\end{table}

\subsection{Qualitive Results}
\begin{figure}
    \centering    \includegraphics[width=1.0\linewidth]{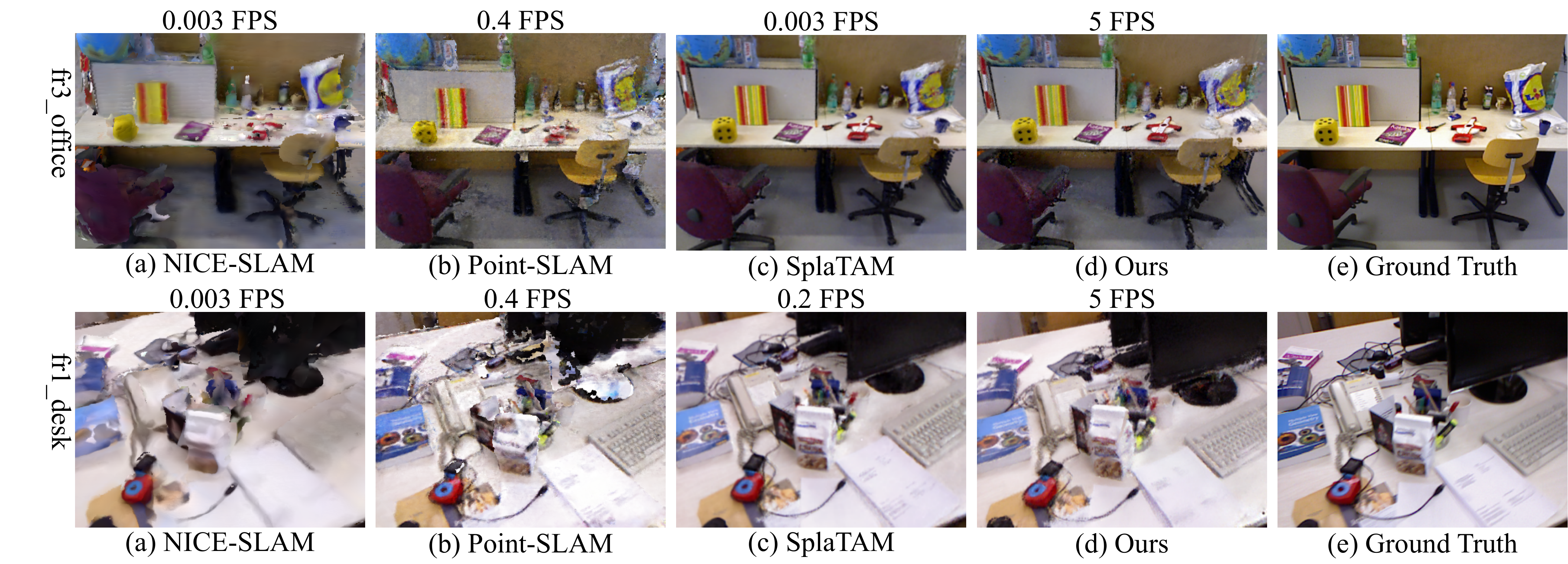}
    \caption{
Qualitative results on TUM RGBD fr1\_desk and fr3\_office sequences.}
    \label{comparison}
\end{figure}

As illustrated in Figure~\ref{comparison}, our method achieves rendering quality that surpasses NeRF-based SLAM and is comparable to SplaTAM, all while maintaining a near-real-time frame rate significantly higher than other approaches, at 5 FPS.
\vspace{-3mm}

\subsection{Ablations}
\label{ablations}
\begin{figure}
    \centering
    \includegraphics[width=1.0\linewidth]{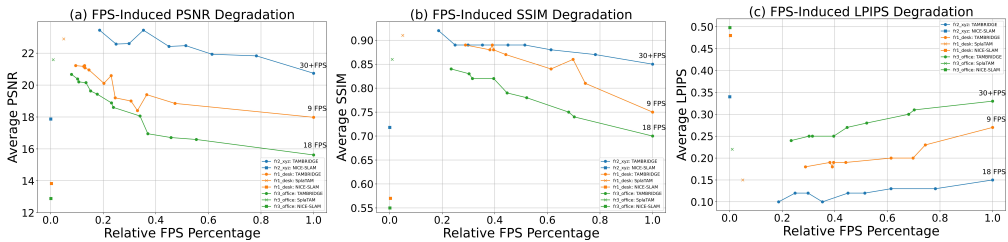}
    \caption{
(a) The change in TAMBRIDGE PSNR with FPS. (b) The change in TAMBRIDGE SSIM with FPS. (c) The change in TAMBRIDGE LPIPS with FPS.}
    \label{fig:FPS}
\end{figure}
\vspace{-2mm}
\textbf{Plug and Play Fusion Bridge.} It is observable that without integrating the modality fusion module, there is a general decline in the reconstruction quality metrics, as shown in Table~\ref{tab:fusion}. This deterioration is attributable to the larger gap between the ORB frontend poses and the Gaussian Splatting, which tends to accumulate over the course of the task. The specific reconstruction results can be seen in the figure below. This highlights one of the most crucial aspects of this work: identifying and demonstrating the significance of the fusion process in bridging traditional SLAM frontend with 3DGS backend. This integration enhances the accuracy and quality of mapping and underscores the necessity of effective pose and data integration techniques in SLAM systems.

\textbf{FPS-Induced Rendering Degradation.}
In robotics perception tasks where real-time performance is crucial, we investigate how rendering quality evolves with increasing frame rates by adjusting the number of optimization iterations \( t \) and the perspective selection threshold \( k \) in the Fusion Bridge module. We offer a detailed set of quality-frame rate curves, as shown in Figure~\ref{fig:FPS}, to facilitate the selection of suitable hyperparameters for specific applications to achieve a balance between real-time performance and rendering quality. Additionally, our approach not only significantly outperforms SplaTAM and NICE-SLAM in terms of frame rate but also maintains a consistent near-real-time performance, consistently exceeding \textbf{5 FPS with best rendering quality}, consistently exceeding \textbf{9 FPS with competent rendering quality} with SOTA NeRF-based methods, and even achieve \textbf{a maximum 30 FPS+ performance} on the scenes with great viewpoint overlappings like fr2/xyz.

%% file: tum_rendering.tex
\begin{table}[t]
    \centering
    \footnotesize
    \tabcolsep=0.08cm 
    \caption{\textbf{Tracking and Rendering Performance on TUM-RGBD~\cite{sturm2012benchmark}.} }
    \resizebox{1.0\textwidth}{!}{
    \begin{tabular}{ccccccccccccc}
    \toprule
    \multicolumn{1}{c}{Common Sequences} & \multicolumn{4}{c}{fr1-desk} & \multicolumn{4}{c}{fr2-xyz} & \multicolumn{4}{c}{fr3-office} \\
    \cmidrule(lr){1-1} \cmidrule(lr){2-5} \cmidrule(lr){6-9}\cmidrule(lr){10-13}
    Method &{\scriptsize RMSE (cm)$\downarrow$} & {\scriptsize PSNR $\uparrow$} &{\scriptsize SSIM $\uparrow$} & {\scriptsize LPIPS $\downarrow$}&{\scriptsize RMSE (cm) $\downarrow$}& {\scriptsize PSNR $\uparrow$} &{\scriptsize SSIM $\uparrow$} & {\scriptsize LPIPS $\downarrow$}&{\scriptsize RMSE (cm) $\downarrow$}& {\scriptsize PSNR $\uparrow$} &{\scriptsize SSIM $\uparrow$ }& {\scriptsize LPIPS $\downarrow$} \\
    \midrule
     ORB-SLAM3 ~\cite{ORBSLAM3_TRO} &{2.06}&-&-&-&0.33&-&-&-&1.04&-&-&-\\ 
    Nice-SLAM~\cite{Zhu2022CVPR}&19.32&12.00&0.42&0.51&36.10&18.20&0.60&0.31&25.31&16.34&0.55&0.39\\ 
    Vox-Fusion~\cite{yang2022vox}&3.52&15.79&0.65&0.52&1.49&16.32&0.71&0.43&26.01&17.27&0.68&0.46\\ 
    Point-SLAM\cite{xu2022point}&4.34&13.87&0.63&0.54&31.73&17.56&0.71&0.59&3.87&18.43&0.75&0.45\\ 
    ESLAM~\cite{johari2023eslam}&3.36&17.49&0.56&0.48&31.44&22.22&0.72&0.23&25.80&19.11&0.61&0.35\\ 
    Co-SLAM~\cite{wang2023coslam}&3.09&16.41&0.48&0.59&31.34&19.17&0.59&0.37&25.37&17.86&0.54&0.45\\ 
    Go-SLAM~\cite{go-slam}&2.11&15.79&0.53&0.53&31.78&16.11&0.53&0.41&26.80&16.49&0.56&0.56\\ SplaTAM~\cite{keetha2023splatam}&3.38&\textbf{22.00}&0.86&0.23&1.34&\textbf{24.50}&\textbf{0.95}&\textbf{0.10}&5.03&\textbf{21.90}&\textbf{0.88}&\textbf{0.20}\\ 
    \textbf{Ours (5 FPS)}&\textbf{1.75}&{21.22}&{\textbf{0.88}}&{\textbf{0.19}}&{\textbf{0.32}}&{23.44}&{0.90}&{\textbf{0.10}}&\textbf{1.42}&20.15&0.82&0.25\\ 
    \textbf{Ours (10 FPS+)}&{\textbf{1.75}}&{17.98}&{0.75}&{0.27}&{\textbf{0.32}}&{20.74}&{0.85}&{0.15}&{\textbf{1.42}}&{16.59}&{0.60}&{0.40}\\ 
    \bottomrule
    \end{tabular}}
\label{tab:tum_rendering}
\end{table}

%% file: tum_mobile.tex
\begin{table}[t]
    \centering
    \footnotesize
        \tabcolsep=0.08cm 
   \caption{\textbf{Tracking and Rendering Performance on TUM-RGBD~\cite{sturm2012benchmark} Mobile Slam Sequence}. Convergence failure is represented by "\xmark" .}
    \resizebox{1.0\textwidth}{!}{
    \begin{tabular}{ccccccccccccc}
    \toprule
	\multicolumn{1}{c}{Mobile SLAM}&  \multicolumn{4}{c}{fr2-pio-360}& \multicolumn{4}{c}{fr2-pio-slam}& \multicolumn{4}{c}{fr2-pio-slam2}\\
 \cmidrule(lr){1-1} \cmidrule(lr){2-5} \cmidrule(lr){6-9}\cmidrule(lr){10-13}
Method &{\scriptsize RMSE (cm) $\downarrow$} &{\scriptsize PSNR $\uparrow$} &{\scriptsize SSIM $\uparrow$} & {\scriptsize LPIPS $\downarrow$}&{\scriptsize RMSE (cm) $\downarrow$}& {\scriptsize PSNR $\uparrow$} &{\scriptsize SSIM $\uparrow$} & {\scriptsize LPIPS $\downarrow$}&{\scriptsize RMSE (cm) $\downarrow$}& {\scriptsize PSNR $\uparrow$} &{\scriptsize SSIM $\uparrow$ }& {\scriptsize LPIPS $\downarrow$} \\
    \midrule
SplaTAM~\cite{keetha2023splatam}&101.37&\textbf{24.33}&\textbf{0.82}&\textbf{0.22}&\xmark&\xmark&\xmark&\xmark&627.31&13.62&0.42&0.54\\
\textbf{Ours}&\textbf{8.43}&{24.32}&{0.79}&{0.24}&{\textbf{10.06}}&{\textbf{23.54}}&{\textbf{0.80}}&{\textbf{0.24}}&\textbf{6.7}&\textbf{24.52}&\textbf{0.82}&\textbf{0.19}\\ 
    \bottomrule
    \end{tabular}
}
    \label{tab:tum_mobile}
\end{table}

%% file: tum_handheld.tex
\begin{table}[!htb]
\centering
 \footnotesize
\tabcolsep=0.08cm 
\caption{\textbf{Tracking and Rendering Performance on TUM-RGBD~\cite{sturm2012benchmark} Handheld Slam Sequence}. Convergence failure is represented by "\xmark" .}
\small
\tabcolsep=1.3mm
\begin{tabular}{lccccccccc}
\toprule
Method & Metric & \text{fr1/360} & \text{fr1/floor} & \text{fr2/hemi}&\text{fr2/kidn}&\text{fr2/desk}&\text{fr2/nloop}&\text{fr2/wloop} &Avg.\\
\midrule
\multirow{4}{*}{\makecell[l]{SplaTAM\textcolor{red}{$^*$}~\cite{keetha2023splatam}}}
& RMSE$\downarrow$ &\textbf{8.64} &  78.9 & \xmark & \xmark & 99.74 & 211.9 & 206.27 & \xmark\\
& PSNR$\uparrow$ & \textbf{18.41} & 18.77 & \xmark & \xmark & 19.26 & 17.89 & \textbf{21.19} &\xmark \\
& SSIM$\uparrow$ &\textbf{0.74} & 0.65 & \xmark &\xmark & 0.77 & \textbf{0.72} & 0.81 &\xmark \\
& LPIPS$\downarrow$ & \textbf{0.27} & 0.38 & \xmark & \xmark & 0.34 & \textbf{0.27} & \textbf{0.21}&\xmark \\ [0.8pt]
\hdashline \noalign{\vskip 1pt}
\multirow{4}{*}{\makecell[l]{\textbf{Ours}}}
& RMSE $\downarrow$ & 10.67 & \textbf{5.98} & \textbf{13.93} & \textbf{8.22} & \textbf{1.02} & \textbf{18.48} & \textbf{1.42} & \textbf{8.53}\\
& PSNR $\uparrow$ & 16.93 & \textbf{20.55} & \textbf{20.63} & \textbf{22.85} & \textbf{20.16} & \textbf{19.49} & 20.15 & \textbf{20.10} \\
& SSIM $\uparrow$ & 0.67 & \textbf{0.71} & \textbf{0.77} & \textbf{0.87} & \textbf{0.8} & \textbf{0.72} & \textbf{0.82} & \textbf{0.77}\\
& LPIPS$\downarrow$ & 0.32 & \textbf{0.27} & \textbf{0.24} & \textbf{0.15} & \textbf{0.21} & 0.29 & 0.25& \textbf{0.25} \\ \noalign{\vskip 1pt}

\bottomrule
\end{tabular}
\label{tab:tum_handheld}
    \vskip -0.2cm
    \vskip -0.1cm
\end{table}    

%% file: 05_conclusion.tex
\section{Conclusion}
Our goal was to develop a 3DGS-based SLAM system suitable for intelligent robotic perception tasks. The system is designed to meet the real-time requirements of robotic perception, and is robust against random sensor noise, motion blur and challenges posed by long-session SLAM. To achieve this, we have implemented a plug-and-play Fusion Bridge module that filters redundant views and leverages the anchoring effect of reprojection errors to estimate poses for rendering initially. This approach marks the first integration of a tracking-centered paradigm with an ORB-based visual odometry frontend together with a 3DGS-centered backend. Our system consistently achieves real-time localization and near-real-time rendering at over 5 FPS on the real-world TUM RGB-D dataset. Furthermore, it significantly surpasses SplaTAM in localization accuracy, rendering quality, and robustness in scenarios involving motion blur and long-distance SLAM tasks.

%% file: appendix.tex
\appendix

\section{Supplementary Material}
\label{app:ei}
\subsection{Implementation Details of TAMBridge}
The detailed implementation of hyper-parameters are shown in Table~\ref{hyper}. $\alpha$ sets the percentage threshold for maximum matching, and $\beta$ sets the minimum numbers of matching points in Viewpoint Selection; $\gamma$ is the threshold of the minimize border mask value of each pixel; $\zeta$ is the weight for SSIM loss and L1 loss in Online 3DGS Mapping; $\tau$ is the maximum threshold of opacity used in Gaussian Pruning; $w_1$, $w_2$, $w_3$ are the weights of reprojection loss, color rendering loss and depth rendering loss, respectively, in Fusion Bridge; $w_4$ and $w_5$ are the weights for depth rendering loss and color rendering loss in Online 3DGS Mapping.
\begin{table}[htbp]
\centering
\caption{Hyper-parameter Values of Implementation}
  \setlength\tabcolsep{2.0mm}
  \label{hyper}
\begin{tabular}{l|cccccccccc}
\toprule
Hyper-parameters &\(\alpha\)&\(\beta\)&\(\gamma\)& \(\zeta\)&\(\tau\) & \(w_1\)&\(w_2\) &\(w_3\)  &\(w_4\)  &\(w_5\)  \\ \hline
  Value & 0.75 & 20 & 0.99 & 0.3&0.005 & 1.5 & 0.5 & 1 & 0.5 & 1  \\ 
\bottomrule
\end{tabular}
\end{table}

\subsection{Evaluation on tiny motion sequences}
The table clearly shows that in sequences with minimal movement, our method outperforms SplaTAM in localization accuracy and matches its reconstruction quality. Due to the presence of many redundant viewpoints in these sequences, where the camera movement is confined to a small area, our selection mechanism achieves superior localization precision and facilitates easier convergence to a comparable level of reconstruction quality. Particularly, in the fr2\_rpy sequence, our method has successfully addressed the convergence issues that previously affected SplaTAM as shown in Table \ref{tab:tum_tiny_motion}.
\input{tum_tiny_motion}
\vspace{-4mm}
\subsection{Reconstruction Quality on TUM RGB-D}
\begin{figure}[htbp]
    \centering    \includegraphics[width=1.0\linewidth]{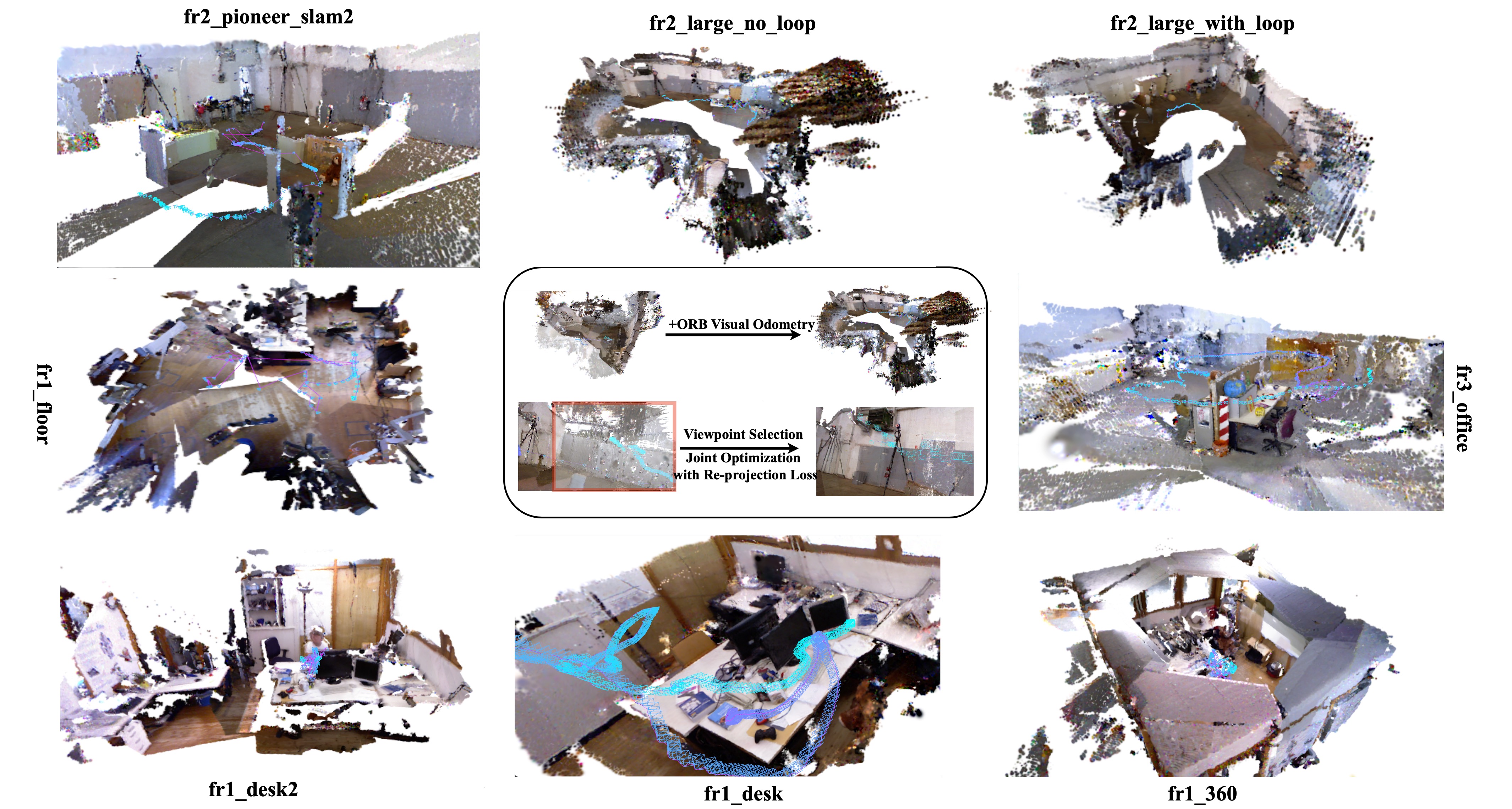}
    \caption{
Reconstruction results on TUM RGBD sequences.}
    \label{s}
\end{figure}

Figure~\ref{s} shows the visualization results on 8 sequences of the TUM RGB-D dataset, where all the perspectives in the images are novel viewpoints, and the sequences contain rich sensor noise and motion blur.
\label{app:metric_scaling:token_edit_distance}

\subsection{FPS Induced Rendering Degradation with Hyper-paramters Values}
\label{FPS}
\begin{figure}[htbp]
    \centering    \includegraphics[width=0.8\linewidth]{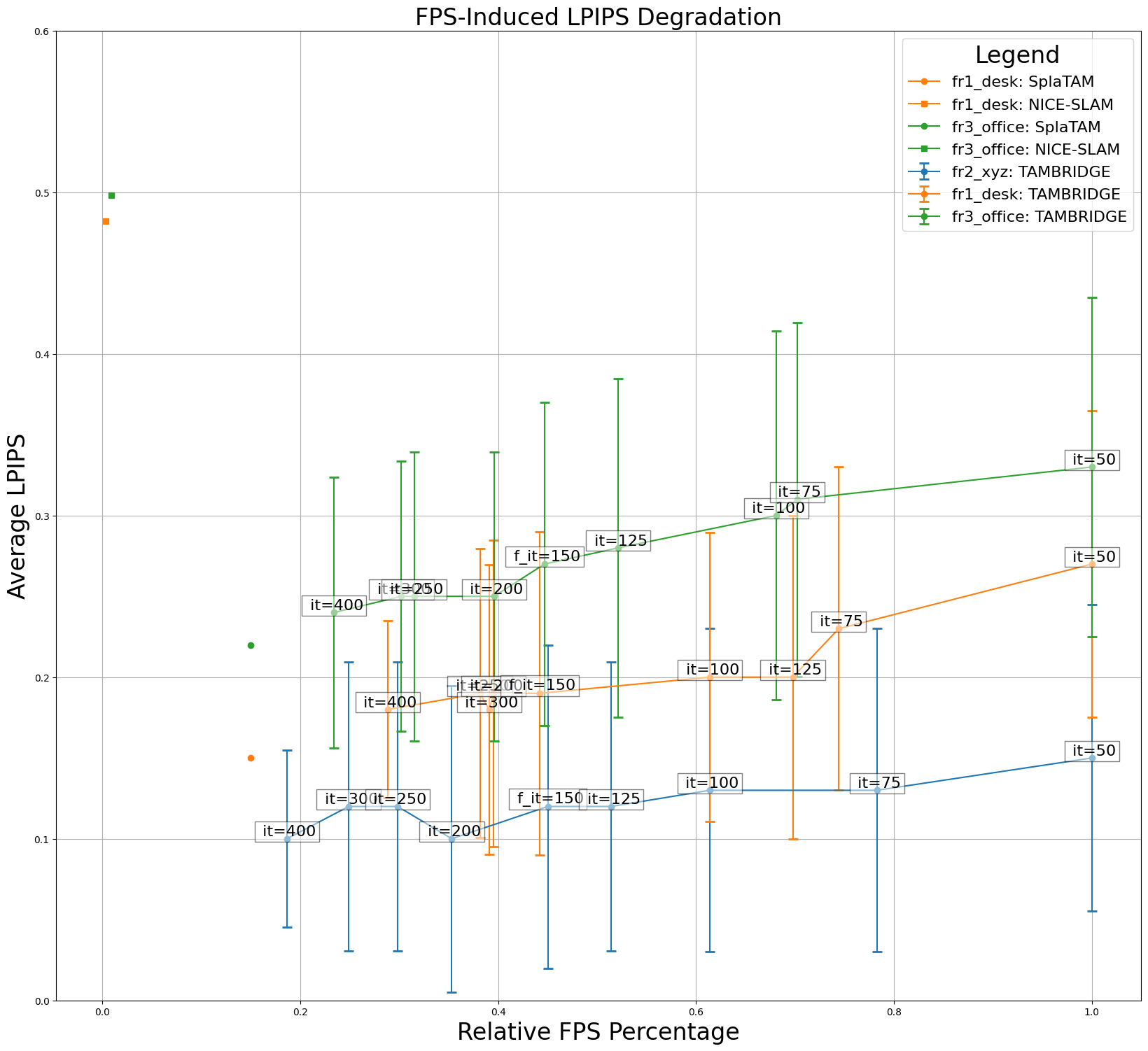}
    \caption{
Reconstruction results on TUM RGBD sequences.}
    \label{17}
\end{figure}

\begin{figure}[htbp]
    \centering    \includegraphics[width=0.8\linewidth]{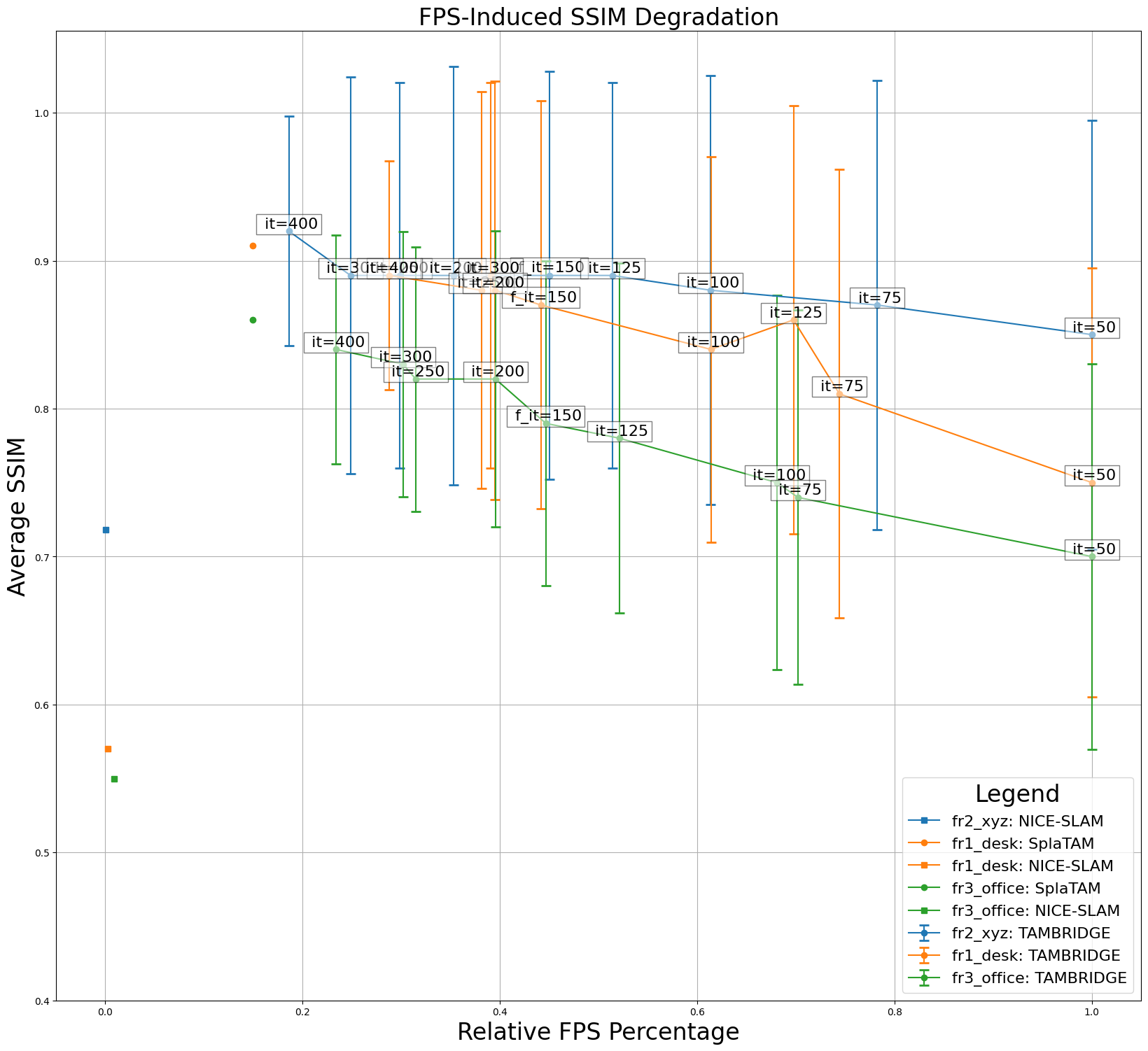}
    \caption{
Reconstruction results on TUM RGBD sequences.}
    \label{18}
\end{figure}

\begin{figure}[htbp]
    \centering    \includegraphics[width=0.8\linewidth]{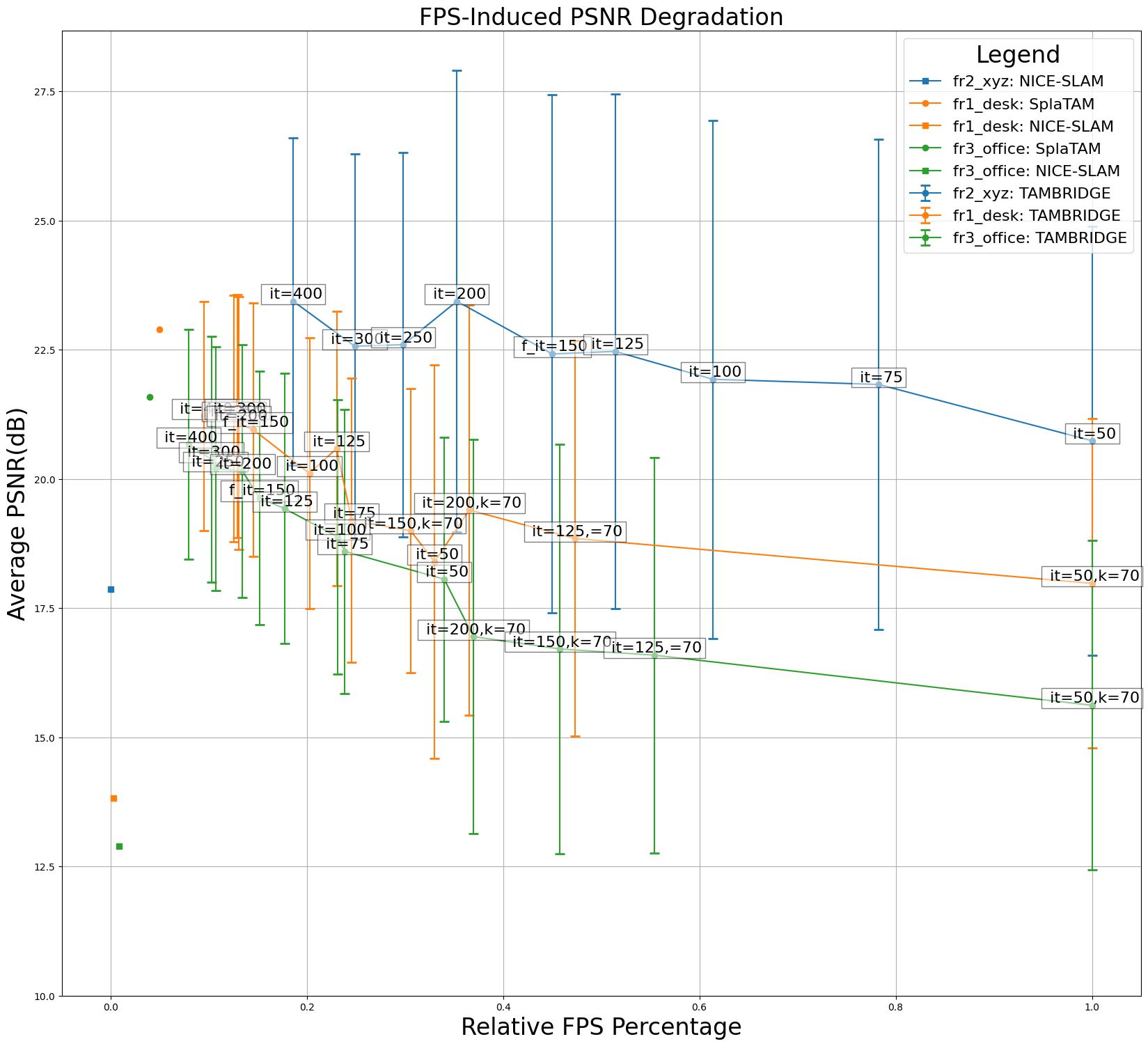}
    \caption{
Reconstruction results on TUM RGBD sequences.}
    \label{19}
\end{figure}

Based on the FPS-Quality~\ref{fig:FPS} curves presented in the experimental section~\ref{experiment}, we have supplemented the data points with corresponding hyperparameter information, as shown in Figure~\ref{17}, Figure~\ref{18}, and Figure~\ref{19}. Specifically, It represents the number of iterations for the joint optimization cycle, and k represents the value of $\alpha$ in the Fusion Bridge module.

\section{Limitations and Future Work}
\label{limi}
\subsection{Limitations of Method}
\label{limi-method}

The limitations of our method are listed as follow:
\begin{enumerate}
    \item The Viewpoint Selection module of the Fusion Bridge, which selects based on the number of keypoint matches and manual thresholds, lacks self-learning capabilities. It can be considered to utilize neural networks such as CNN or Transformer to learn autonomously and determine the degree of viewpoint overlap. Additionally, other factors like sensor noise can also be introduced to filter the original image quality.
    \item After the viewpoint selection, the reconstructed map may contain some voids, which, although having no practical impact on scene understanding and usage, need to be filled with effective methods or avoided during the selection process.
    \item The online 3DGS module builds upon existing 3DGS-based SLAM methods. However, many lightweight improvements to 3DGS have been proposed in the field of 3D reconstruction. Adapting these new 3DGS methods for online reconstruction could further enhance the real-time performance and effectiveness of the current approach.
\end{enumerate}
\subsection{Limitations of Experiment}
\label{limi-experiment}
\begin{figure}[htbp]
    \centering    \includegraphics[width=1.0\linewidth]{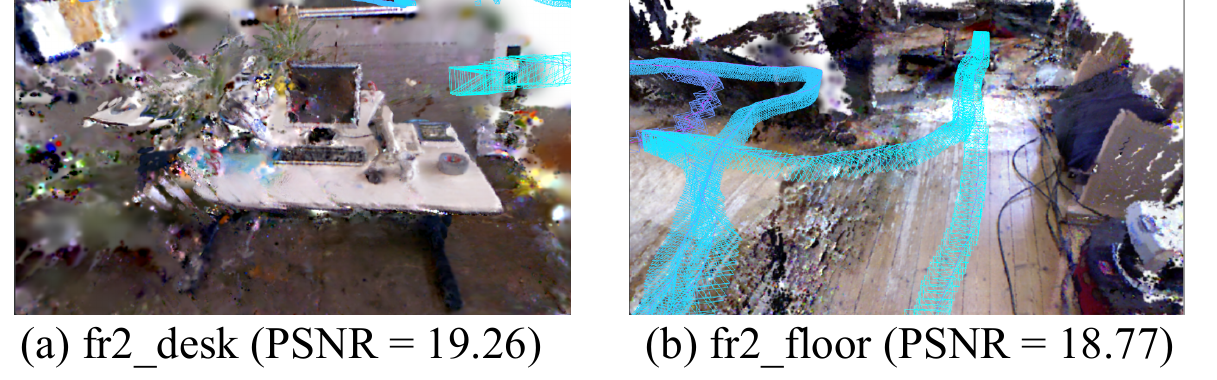}
    \caption{
Noisy scenes rendered by SplaTAM.}
    \label{noise}
\end{figure}

\begin{enumerate}
    \item \textbf{Metrics.} SplaTAM and other 3DGS-based SLAM are commonly evaluated using metrics such as PSNR, LPIPS, and SSIM. However, most of these metrics calculate their values by averaging each frame in the sequence, which does not accurately assess the overall quality of the environmental reconstruction. For instance, in experiments, SplaTAM may exhibit good single-frame indicators in some long-distance sequences, but due to the lack of a global evaluation metric, the reconstructed scene may contain significant noise and viewpoint-overlapping, even to the point of being unintelligible, as shown in Figure~\ref{noise}.
    \item \textbf{Frontends.} Additional experiments can be conducted with other traditional SLAM front-ends such as VINS-MONO. The plug-and-play nature of the Fusion Bridge allows it to connect with any sparse feature-based front-end and an online 3DGS back-end.
\end{enumerate}
\begin{table}[h]
    \centering
    \footnotesize
    \setlength{\tabcolsep}{5.8mm}
    \caption{\textbf{Experimental Results without Fusion Bridge.} }
    \begin{tabular}{cccc}
    \toprule
\multirow{2}{*}{Method} &\multicolumn{3}{c}{HRI\_MEETING}\\
\cmidrule(lr){2-4}
 &{\scriptsize PSNR $\uparrow$} &{\scriptsize SSIM $\uparrow$} & {\scriptsize LPIPS $\downarrow$}\\
    \midrule
\textbf{SplaTAM}&16.93&\textbf{0.62}&0.41\\  \textbf{Ours}&{\textbf{17.91}}&{\textbf{0.62}}&{\textbf{0.38}}\\ 
    \bottomrule
    \end{tabular}
    \label{tab:hri}
\end{table}
\section{Appendix Experiment on HRI\_MEETING Sequence}
\begin{figure}[h]
    \centering    \includegraphics[width=0.3\linewidth]{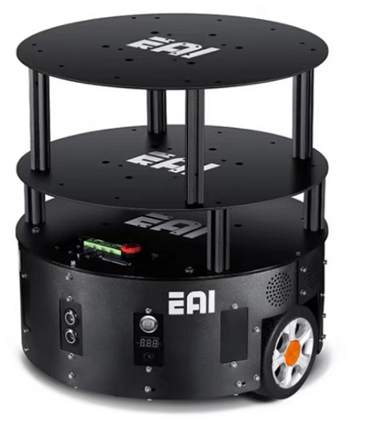}
\caption{The Dashgo-E1 mobile robot.}
    \label{dashgo}
\end{figure}

We conducted real-world experiments in our lab's conference room using a Dashgo-E1 mobile robot as shown in Figure~\ref{dashgo} equipped with a Realsense-d435i RGB-D camera. The experimental results, as shown in the table~\ref{tab:hri}, indicate that our method outperforms SplaTAM in both reconstruction quality and localization accuracy. More sequences will be added and published as a new dataset in the future.

\label{lab}

%% file: tum_tiny_motion.tex
\begin{table}[!htb]
\centering
\small
\tabcolsep=5.4mm
\caption{\textbf{Tracking and Rendering Performance on TUM-RGBD~\cite{sturm2012benchmark} Tiny Motion Sequence}. Convergence failure is represented by "\xmark" .}
\begin{tabular}{lcccccc}
\toprule
Method & Metric & \texttt{fr1/xyz} & \texttt{fr1/rpy} & \texttt{fr2/rpy} & Avg.\\
\midrule
\multirow{4}{*}{\makecell[l]{ORB-SLAM3~\cite{ORBSLAM3_TRO}}}
& RMSE$\downarrow$ & \textbf{1.02} & 1.9 & 0.46 & 1.13 \\
& PSNR$\uparrow$ & -- & -- & --  & --\\
& SSIM$\uparrow$ & -- & -- & --  & --\\
& LPIPS$\downarrow$ & -- & -- & -- & --\\  \hdashline
\multirow{4}{*}{\makecell[l]{SplaTAM\textcolor{red}{$^*$}~\cite{keetha2023splatam}}}
& RMSE$\downarrow$ &3.38 & \xmark & 5.03 & \xmark\\
& PSNR$\uparrow$ & \textbf{22.89} & \xmark & \textbf{21.29} & \xmark \\
& SSIM$\uparrow$ &\textbf{ 0.91} & \xmark &\textbf{0.86} & \xmark \\
& LPIPS$\downarrow$ & \textbf{0.15 }& \xmark & \textbf{0.22} & \xmark \\ [0.8pt] \hdashline \noalign{\vskip 1pt}
\multirow{4}{*}{\makecell[l]{\textbf{Ours}}}
& RMSE$\downarrow$ &\textbf{1.29} &  \textbf{0.32} & \textbf{1.42} & \textbf{1.01}\\
& PSNR$\uparrow$ & 21.08 & \textbf{23.16} &20.15 & \textbf{22.33} \\
& SSIM$\uparrow$ &0.88 & \textbf{0.91} & 0.82 & \textbf{0.87} \\
& LPIPS$\downarrow$& 0.19 &\textbf{0.10} & 0.25& \textbf{0.19} \\[0.8pt]  \noalign{\vskip 1pt}
\bottomrule
\end{tabular}
\label{tab:tum_tiny_motion}
\end{table}